\documentclass{article} 
\usepackage{nips14submit_e,times}
\usepackage{hyperref}
\usepackage{url}

\title{Hierarchical models for neural population dynamics in the presence of non-stationarity} %

\author{
Mijung Park \\
Gatsby Computational Neuroscience Unit, University College London \\
Max Planck Institute for Biological Cybernetics T{\"u}bingen \&  \\
Bernstein Center for Computational Neuroscience T{\"u}bingen\\ 
\texttt{mijung@gatsby.ucl.ac.uk} 
\And
Jakob H. Macke \\
Max Planck Institute for Biological Cybernetics T{\"u}bingen\\
Bernstein Center for Computational Neuroscience T{\"u}bingen\\
\texttt{jakob@tuebingen.mpg.de} \\
}

\usepackage{amsthm}
\usepackage{algorithm,algorithmic}
\usepackage{wrapfig} 
\usepackage{sidecap}  
\usepackage{amsmath} 
\usepackage{amssymb} 
\usepackage{amsfonts}
\usepackage{bm,amsbsy} 
\usepackage{graphicx}
\usepackage{cite} 
\usepackage{color}  
\usepackage[labelfont=bf,labelsep=period,justification=raggedright]{caption}
\usepackage{wrapfig} 
\usepackage{sidecap}  
\usepackage[small,tight]{bibhacks} 

\newcommand{\Nrm}{\mathcal{N}}
\newcommand{\Dat}{\mathcal{D}}

\newcommand{\vx}{\mathbf{x}}

\newcommand{\vu}{\mathbf{u}}
\newcommand{\vy}{\mathbf{y}}

\newcommand{\trp}{{^\top}} 

\newcommand{\va}{\mathbf{a}}
\newcommand{\vb}{\mathbf{b}}
\newcommand{\vd}{\mathbf{d}}   
 
\newcommand{\vh}{\mathbf{h}} 
\newcommand{\vc}{\mathbf{c}}

\newcommand{\vmu}{\mathbf{\ensuremath{\bm{\mu}}}}

\newcommand{\vtheta}{\mathbf{\ensuremath{\bm{\theta}}}}

\newcommand{\vphi}{\mathbf{\ensuremath{\bm{\phi}}}}

\renewcommand{\eqref}[1]{eq.~\ref{eq:#1}}
\newcommand{\figref}[1]{Fig.~\ref{fig:#1}}

\nipsfinalcopy 

\begin{document}

\maketitle

\begin{abstract}

Neural population activity often exhibits  rich variability and temporal structure. This variability is thought to arise from single-neuron stochasticity, neural dynamics on short time-scales, as well as from modulations of neural firing 
properties on long time-scales, often referred to as ``non-stationarity". 
To better understand the nature of co-variability in neural circuits and their impact on cortical information processing, we need statistical models that are able to capture multiple 
sources of variability on different time-scales. 
Here, we introduce a hierarchical statistical model of neural population activity which models both neural population dynamics as well as inter-trial modulations in firing rates.
In addition, we extend the model to allow us to capture non-stationarities in the population dynamics itself (i.e., correlations across neurons).
 We develop variational inference methods for learning model parameters, and demonstrate that the method can recover non-stationarities in both average firing rates and correlation structure.
Applied to neural population recordings from anesthetized macaque primary visual cortex, our models provide a better account of the structure of neural firing than stationary dynamics models. 
\end{abstract}

\section{Introduction}

Neural spiking activity recorded from populations of cortical neurons exhibits substantial variability in response to repeated presentations of a sensory stimulus 
\cite{Renart_Machens_14}. This variability is thought to arise both  
from dynamics generated endogenously within the circuit \cite{Destexhe_11}, as well as from internal and behavioural states  \cite{Maimon_11,Harris_Thiele_11, 
Ecker_Berens_14}. An understanding of how the 
interplay between sensory inputs and endogenous dynamics shapes neural activity patterns
 is essential for our understanding of how information is processed by 
neuronal populations. Multiple 
statistical \cite{Smith_Brown_03,Eden_Frank_04,Yu_Afshar_06, Kulkarni_Paninski_07, Truccolo_Hochberg_10, Macke_Buesing_11} and mechanistic 
\cite{Vreeswijk_Sompolinsky_96} models for 
characterising neuronal population dynamics have been developed. 

In addition to these dynamics which take place on fast time-scales (milliseconds up to few seconds), there are also processes modulating neural firing activity which take place 
on much slower time-scales 
(seconds to days).  The slow drifts in rates across an experiment can be caused by fluctuations in arousal, anaesthesia level or physiological properties of the experimental 
preparation \cite{Tomko_Crapper_74, Brody_99,Goris_Movshon_14}. Furthermore, processes such as learning and short-term plasticity can lead to changes in neural firing properties 
\cite{Gilbert_Li_12}.  The statistical structure of these slow fluctuations has been modelled using state-space models and related techniques 
\cite{Brown_Nguyen_01,Frank_Eden_02,Lesica_Stanley_05,Ventura_Cai_05, 
Quiroga-Lombard_Hass_13}.

To accurately capture the the structure of neural dynamics and to disentangle the contributions of slow and fast modulatory processes to neural variability and co-variability, 
it is therefore important to develop models 
that can capture neural dynamics both on fast (i.e.,  within experimental trials) and slow (i.e., across trials) time-scales. Czanner et al. \cite{Czanner_Eden_08} presented a 
statistical model of single-neuron 
firing  in which within-trial dynamics are modelled by (generalised) linear coupling from the recent spiking history of each neuron onto its instantaneous firing rate,
 and across-trial dynamics were modelled by defining a random walk model over parameters.   More recently, Mangion et al \cite{Mangion_Yuan_11} presented a latent linear 
dynamical system model with Poisson observations (PLDS, \cite{Smith_Brown_03,Kulkarni_Paninski_07,Macke_Buesing_11}) with a one-dimensional latent space, and used  a 
heuristic filtering-approach for tracking 
parameters, again based on a random-walk model.

Here, we present a hierarchical Bayesian model of neural dynamics. Our model consists of a latent dynamical system with Poisson observations (PLDS) to model neural 
population dynamics combined with a Gaussian process (GP)  \cite{Rasmussen_Williams_06} 
to model modulations in firing rates or model-parameters across experimental trials. 
Compared to the random-walk based prior work, using a GP is a more flexible and powerful way of defining the distributions over the non-stationarity, with hyperparameters that control its variability and smoothness. 

We focus on two concrete variants of this general model: In the first variant, we introduce a new set of variables which control neural firing rate on each trial to capture non-stationarity in firing rates. In the second variant, we allow the 
dynamics matrix which determines the spatio-temporal correlations in the population to vary across trials.
We derive a variational Bayesian (VB) inference method for estimating the posterior distribution over (possibly time-varying) parameters from population recordings of 
spiking activity. 
Our approach generalises the $1$-dimensional latent states in \cite{Mangion_Yuan_11} to models with multi-dimensional states, as well as to a Bayesian treatment of non-stationarity based on Gaussian Process priors. 

\section{Hierarchical non-stationary models of neural population dynamics}



We start by introducing a hierarchical model for capturing short time-scale population dynamics as well as long time-scale non-stationarities in firing rates or model parameters. Although 
we use the term ``non-stationary" to mean that the system is 
best described by parameters that change over time, the distribution over parameters can be described by a stochastic process which might be strictly stationary in the 
statistical sense\footnote{A stochastic process is strict-sense stationary if its joint distribution over  any two time-points $t$ and $s$ only depends on the elapsed time $t-s$.}.

\paragraph{Modelling framework.} We assume that the neural population activity $\vy_t$ depends on a $k$-dimensional latent state $\vx_t \in \mathbb{R}^{k}$ and a modulatory factor $\vh^{(i)}
\in \mathbb{R}^k $ which is different for each trial $i = \{1, \ldots, r \}$.
 The latent state $\vx$ models short-term co-variability of spiking activity and the modulatory factor $\vh$ models slowly varying mean firing rates across experimental trials.
 
We model neural spiking activity of $p$ neurons as conditionally Poisson given the latent state $\vx_t$ and a modulator $\vh^{(i)}$, with a 
log firing rate which is linear in parameters and latent factors, 
\begin{align}\label{eq:Likeli_NSPLDS}
\vy_t|\vx_t, C, \vh^{(i)}, \vd &\sim  \mbox{Poiss}(\vy_t | \exp(C (\vx_t + \vh^{(i)}) +  \vd )),
\end{align}
where the loading matrix $C \in \mathbb{R}^{p \times k}$ specifies how each neuron is related to the latent state and the modulator, and  $\vd  \in \mathbb{R}^{p}$ is an offset term that 
controls the mean 
firing rate of each cell.  We note that, because of the use of an exponential firing-rate nonlinearity, latent factors have a multiplicative effect on neural firing rates, as has been 
observed experimentally 
\cite{Goris_Movshon_14, Ecker_Berens_14}.

Following \cite{Kulkarni_Paninski_07,Macke_Buesing_11,Mangion_Yuan_11}, we assume that the latent dynamics evolve according to a first-order autoregressive process with 
Gaussian innovations, 
\begin{align}\label{eq:LDS}
\vx_t|\vx_{t-1}, A, B, Q &\sim \Nrm(\vx_t | A\vx_{t-1} + B \vu_t, Q).
\end{align}
Here, we allow for sensory stimuli (or experimental covariates), $\vu_t \in \mathbb{R}^d$ to influence  the latent states linearly.  The dynamics matrix $A \in \mathbb{R}^{k \times k}$ determines the 
state evolution, $B \in 
\mathbb{R}^{k \times d}$ models the dependence of latent states on external inputs, and $Q\in \mathbb{R}^{k \times k}$ is the noise covariance $Q\in \mathbb{R}^{k \times k}$. 
We set $Q$ to be the identity matrix, $Q= \mathbf{I}_k$ as in \cite{Beal_03}, and we assume $\vx_0^{(i)} \sim \Nrm(0, \mathbf{I}_k)$. 

The parameters in this model are $\vtheta=\{A, B, C, \vd, \vh^{(1:r)} \}$. 
We refer to this general model as {\it{non-stationary PLDS}} (N-PLDS). 
Different variants of N-PLDS can be constructed by placing priors on individual parameters which allow them to vary across trials (in which case they would then 
depend on the trial index $i$) or by omitting different components of the model. In the following, we will focus on two concrete variants.

\begin{figure}[t]
\includegraphics[width=1\linewidth]{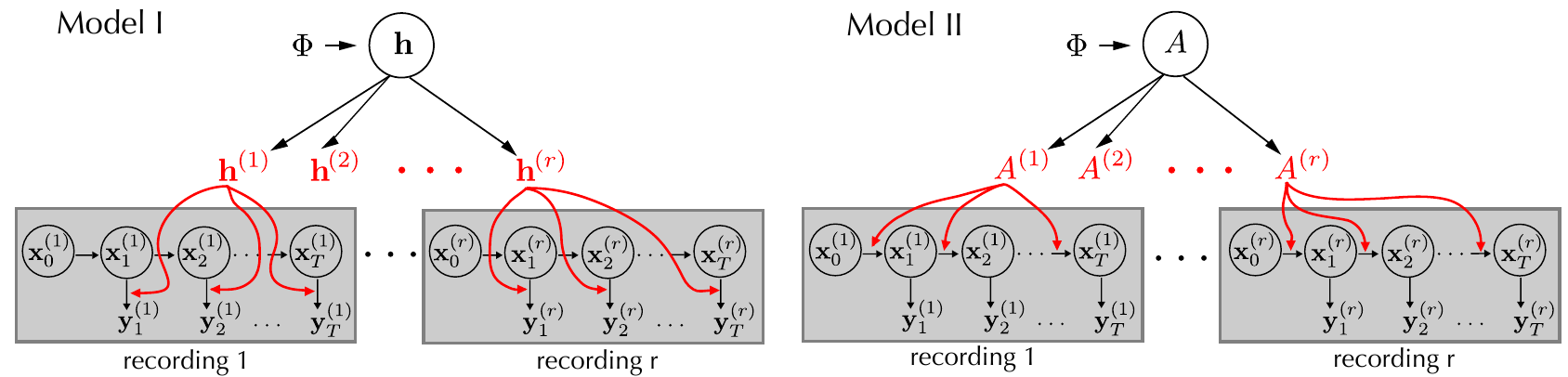} 
\caption{Schematic of hierarchical non-stationary Poisson observation Latent Dynamical System (N-PLDS) models. \textbf{Model I} for capturing non-stationarity in mean firing 
rates. The parameter $\vh$ slowly varies across trials and leads to fluctuations in mean firing rates. \textbf{Model II} for capturing non-stationarity in population dynamics. The 
dynamics matrix $A$ changes across trials, as controlled by the hyperparameters $\Phi$.}
\label{fig:Schematic}
\end{figure}

\paragraph{Model  I : Non-stationarity in mean firing rates.} 
In the first model (\figref{Schematic} Left), we assume all  parameters apart from $\vh$ to be constant across 
trials. For the modulator $\vh$, we assume that it varies across trials according to a GP with mean $0$ and (modified) squared exponential kernel, i.e. 
$\vh^{(i)} \sim \mathcal{GP}(0, K(i, j))$, where the  $(i, j)$th block of $K$ (size $k\times k$) is given by
\begin{align}\label{eq:compact_cov_priors_A}
K(i, j) =  (\sigma^2 + \epsilon \delta_{i,j}) \exp \left(-\tfrac{1}{2 \tau^2} (i-j)^2  \right)  I_{k}.
\end{align} 
Here, we assume the independent noise-variance on the diagonal ($\epsilon$)  to be constant and small as in \cite{Yu_Cunningham_09}. When $\sigma^2 =\epsilon=0$, the modulator 
vanishes, which corresponds to the conventional PLDS model with fixed parameters \cite{Kulkarni_Paninski_07,Macke_Buesing_11}.  When $\sigma^2 >0$, the mean firing rates vary across trials, and the parameter $\tau$ determines the time-scale (in units of `trials') of these fluctuations.
 We impose ridge priors on the model parameters (see Appendix A for details), so that the total set of hyperparameters of the model is $\Phi = \{  \sigma^2, \tau^2,  \vphi \}$, where $
\vphi$ is the set of ridge parameters. 

\paragraph{Model II : Non-stationarity in population dynamics.} With the second variant (\figref{Schematic} Right), we aim to capture non-stationarity in dynamics. We drop the modulator $\vh$, and instead allow the dynamics matrix $A$ to vary across trials. That is, we allow the model to have a  different matrix $A^{(i)}$ for each trial $i$.   We 
assume that the evolution of A's over time follows a Gaussian process, 
$\va^{(i)} \sim \mathcal{GP}(\bar{\va}, K(i, j))$, defined over the vectorized matrices $\va^{(i)} = \mbox{vec}(A^{(i)}\trp) \in \mathbb{R}^{k^2}$. Here, $\bar{\va}$ is the (vectorized) 
mean dynamics matrix, the  covariance kernel $K$ is as for model $I$, with the only exception that the blocks are now of size $k^2 \times k^2$ as each $\va^{(i)}$ has $k^2$ 
elements. As before, when $\sigma^2=\epsilon=0$, this model reduces to a stationary PLDS.  When $\tau$ goes to zero,
this model corresponds to  independently drawing $\va^{(i)}$ from $\Nrm(\bar{\va}, \sigma^2 
I_{k^2})$, i.e. a model in which the dynamics are slightly different on each trial, but the fluctuations are not correlated across trials.  For larger values of $\tau$, the squared 
exponential kernel ensures that $A$ will vary smoothly across trials. The parameters in this model are $\vtheta = \{A^{(1:r)}, B, C, \vd \}$. We also impose ridge priors on $\{ B, 
C, \vd \}$, which results in hyperparameters $\Phi = \{  \sigma^2, \tau^2,  \vphi\}$.

\section{Variational Bayesian Expectation Maximization}

Our goal is to infer parameters and latent variables in each model. The exact posterior distribution in each model is not 
analytically tractable due to the Poisson likelihood 
term. Here, we approximate the posterior over the parameters and the latent variables by maximising the lower bound of the marginal likelihood of the observations, 
\begin{eqnarray}\label{eq:variationalLB}
\log p(\vy^{(1:r)}_{1:T}) \geq \int d\vtheta \; d\vx^{(1:r)}_{1:T} \; q(\vtheta,\vx^{(1:r)}_{1:T}) \; \log \frac{p(\vtheta, \vx^{(1:r)}_{1:T}, \vy^{(1:r)}_{1:T})}{q(\vtheta, \vx^{(1:r)}_{1:T})},
\end{eqnarray} where the approximate posterior is denoted by 
$q(\vtheta, \vx^{(1:r)}_{1:T}) = q_{\vtheta}(\vtheta) \prod_{i=1}^r q_{\vx}(\vx^{(i)}_{0:T}), $
which we assume to be factorized between latents and parameters.  We maximize the lower bound by iterating the variational Bayesian expectation maximization (VBEM) algorithm \cite{Beal_03}, which 
consists of (1) the variational Bayesian 
expectation (VBE) step for computing $q_{\vx}(\vx^{(1:r)}_{0:T})$, 
\begin{equation}\label{eq:Estep_Mbeal}
q_{\vx}(\vx^{(1:r)}_{0:T}) \propto \exp \left[ \int d \vtheta q_\vtheta(\vtheta) \log p( \vx^{(1:r)}_{1:T}, \vy^{(1:r)}_{1:T} |\vtheta) \right], 
\end{equation}
and (2) the variational Bayesian maximization (VBM) step  for computing $q_{\vtheta}(\vtheta)$,
\begin{equation}\label{eq:Mstep_Mbeal}
q_\vtheta(\vtheta) \propto p(\vtheta) \exp \left[ \int d \vx^{(1:r)}_{0:T} q_\vx(\vx^{(1:r)}_{0:T}) \log p( \vx^{(1:r)}_{0:T}, \vy^{(1:r)}_{1:T} |\vtheta) \right].
\end{equation} While a complete derivation of the algorithm is presented in Appendix A, we here highlight its main steps.

\paragraph*{VBE step.}
Using the first-order dependency in latent states, we derive a sequential forward/backward algorithm to obtain $q_\vx(\vx^{(1:r)}_{0:T})$, generalising the approach of 
\cite{Mangion_Yuan_11} to multi-dimensional latent states.  Since the VBE step decouples across trials, this step is easy to parallelise across trials (and is not more expensive than Bayesian inference for a `fixed parameter' PLDS) and we omit the trial-indices 
for clarity.
The forward message $\alpha(\vx_t) $ at time $t$ is given by  
\begin{equation}\label{eq:forward}
\alpha(\vx_t) \propto \int d\vx_{t-1} \alpha(\vx_{t-1}) \exp \left[ \langle \log (p(\vx_t|\vx_{t-1})p(\vy_t|\vx_t)) \rangle _{q_{\vtheta}(\vtheta) } \right].
\end{equation} 
Assuming that the forward message at time $t-1$ denoted by $\alpha(\vx_{t-1})$ is Gaussian, the Poisson likelihood term will render the forward message at time 
$t$ non-Gaussian, but we will approximate $\alpha(\vx_t)$ as a Gaussian using the first and second derivatives of the right-hand side of \eqref{forward}. \footnote{This approximation does not guarantee monotonically increasing lower bound after each VBEM iteration. However, we monitored its convergence by computing one-step ahead prediction scores and found that the approximation yielded accurate results as shown in Applications.  }

Similarly,  the backward message at time $t-1$ is given by 
\begin{equation}\label{eq:backward_t1}
\beta(\vx_{t-1}) \propto \int d \vx_t \beta(\vx_t) \exp \left(\langle \log(p(\vx_t|\vx_{t-1})p(\vy_t|\vx_t)) \rangle_{q_{\vtheta}(\vtheta)} \right), 
\end{equation} 
which we also approximate to a Gaussian for tractability in computing backward messages. 

Using the forward/backward messages, we compute the posterior marginal distribution over latent variables (See Appendix A). 
We also need to 
compute the cross-covariance between neighbouring latent variables to obtain the sufficient statistics of latent variables (which we will need for VBM step). 
The pairwise marginals of latent variables are given by
\begin{eqnarray}\label{eq:marg_xt_xt1}
 p(\vx_t, \vx_{t+1}|\vy_{1:T}) 
&\propto& \beta(\vx_{t+1}) \exp \left(\langle \log(p(\vy_{t+1}|\vx_{t+1}) p(\vx_{t+1}|\vx_t)) \rangle_{q_\vtheta(\vtheta)} \right) \alpha(\vx_{t}),
\end{eqnarray} which we approximate as a jointly Gaussian distribution by using the first/second derivatives of \eqref{marg_xt_xt1} and extracting the cross-covariance term from 
the joint covariance matrix.

\paragraph*{VBM step.} In the VBM step, we need to update the posterior distribution over the parameters.  Under Model I, the posterior over parameters factorizes as
\begin{eqnarray}\label{eq:approximate_posterior_params_Model1}
q_{\vtheta}(\vtheta) &=& q_{\va, \vb}(\va, \vb) \; q_{\vc, \vd, \vh}(\vc, \vd, \vh^{(1:r)}),
\end{eqnarray} where vectorized notations $\vb = \mbox{vec}(B\trp)$ and $\vc = \mbox{vec}(C\trp)$. For this model, we set $\vc, \vd$ to the maximum likelihood estimates $\hat\vc, \hat\vd$ for 
simplicity in inference.  The computational cost of this algorithm is dominated by the cost of calculating the posterior distribution over $\vh^{(1:r)}$, which involves manipulation of a $rk$-dimensional Gaussian. While this was still tractable without further approximations for the data-set sizes used in our analyses below, we note that a variety of approximate methods for GP-inference exist which could be used to improve efficiency of this computation. In particular, we will typically be dealing with systems in which $\tau \gg1$, which means that the kernel-matrix is smooth and can be approximated using low-rank representations \cite{Rasmussen_Williams_06}.

Under Model II, the posterior over $\va$ is replaced by $\va^{(1:r)}$, and $\vh^{(1:r)}$ is omitted from \eqref{approximate_posterior_params_Model1}. We then approximate each 
factor in \eqref{approximate_posterior_params_Model1} to a multivariate normal distribution using the first/second derivatives of \eqref{Mstep_Mbeal} w.r.t. each of the 
parameters. Again, computation time is dominated be the GP term, which however lends itself well to low-rank approximations.

\paragraph*{Hyperparameter estimation.}
In each iteration of VBEM algorithm, we also update the hyperparameters by maximizing the lower bound w.r.t. hyperparameters. Given the approximate posterior $q(\vtheta, \vx^{(1:r)}_{0:T}) $, the 
variational lower bound simplifies to (see  Ch.5 in \cite{Beal_03} for details)
\begin{eqnarray}\label{eq:variationalLB_calculation}
\log p(\vy^{(1:r)}_{1:T}|\Phi) \geq - KL(\Phi) + c.
\end{eqnarray} 
The KL divergence between the prior and posterior over parameters, denoted by  $\Nrm(\vmu_{\Phi}, \Sigma_{\Phi})$ and $\Nrm(\vmu, \Sigma)$, respectively, is given by 
\begin{eqnarray}\label{eq:KLAB}
KL(\Phi) &=
& -\tfrac{1}{2} \log| \Sigma_{\Phi}^{-1} \Sigma | + \tfrac{1}{2} \text{Tr} \left[ \Sigma_{\Phi}^{-1} \Sigma \right] +  \tfrac{1}{2} (\vmu - \vmu_{\Phi})\trp \Sigma_{\Phi}^{-1}(\vmu - \vmu_{\Phi})  + c,
\end{eqnarray} where the prior mean and covariance depend on the hyperparameters.

\paragraph*{Predictive distributions for test data.}
In our model, different trials are no longer considered as independent, so we can predict parameters for held-out trials. Using the GP model on $\vh$ and $\va$, we have closed 
form predictive distributions on $\vh^*$ and $\va^*$ for test data $\Dat^*$ given training data $\Dat$: 
\begin{eqnarray}
p(\vh^*|\Dat, \Dat^*) &=& \Nrm(K^* K^{-1} \vmu_{\vh}, \; K^{**} - K^* (K+H_{\vh}^{-1})^{-1} K^*\trp),\\
p(\va^*|\Dat, \Dat^*) &=& \Nrm(\bar{\va} + K^* K^{-1} (\vmu_{\va} - \bar{\va}), K^{**} - K^* (K+H_{\va}^{-1})^{-1} K^*\trp), 
\end{eqnarray} where  $K$ is the prior covariance matrix on $\Dat$ and $K^{**}$ is on $ \Dat^*$, and $K^{*}$ is their prior cross-covariance as introduced in Ch.2 of \cite{Rasmussen_Williams_06}, 
and the negative Hessians $H_\vh$ and $H_\va$ are defined in Appendix A.  

\section{Applications}

\begin{figure}[t]
{\centering
\includegraphics[width=0.9\linewidth]{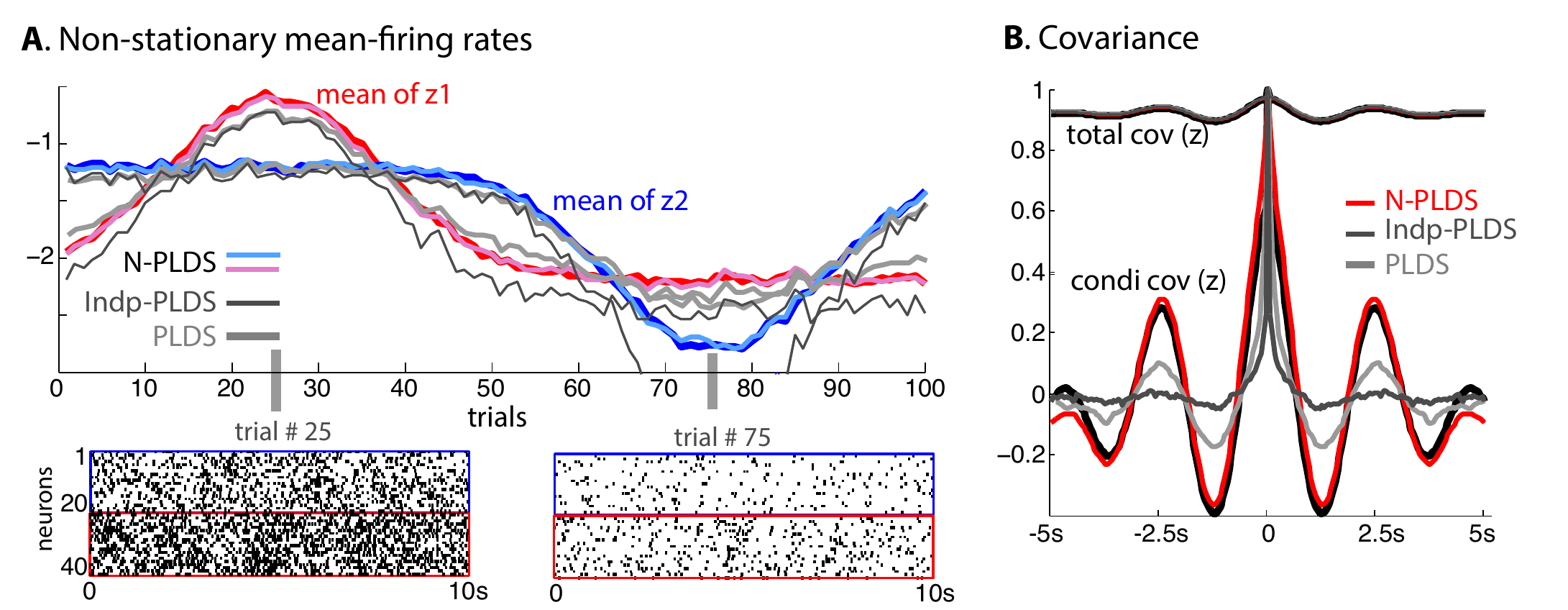} }
\caption{\textbf{Illustration of non-stationarity in firing rates (data simulated from Model I)}. \textbf{A}: Spike rates of $40$ neurons are influenced by two slowly varying firing rate modulators. The log mean firing rates of the two groups of neurons are $z_1$(red, group 1) and $z_2$(blue, group 2) across $100$ trials (top). Raster plots (bottom) show the extreme cases, i.e. trials $25$ and $75$. The traces show the 
posterior mean of $z$ estimated by N-PLDS (Model I, light blue for $z_2$, light red for $z_1$), independent PLDSs (fit a PLDS to each trial data individually, dark gray), and PLDS (light gray). \textbf{B}: Total and 
conditional (on each trial) covariance of recovered neural responses from each model (averaged across all neuron pairs, and then normalised for visualisation).  The covariances recovered by our model (red) well match the true ones (black), while 
those by independent PLDSs (gray) and a single PLDS (light gray) do not.}
\label{fig:non-stationaryFR_simulatedData}
\end{figure}

\subsection{Simulated data} 

We first illustrate the performance of Model I 
 on a simulated (from Model I) population recording from $40$ neurons consisting of  $100$ trials of length $T=200$ time steps each.  We used a  $4$-dimensional latent state  and assumed 
that the population consisted of two homogeneous sub-populations of size $20$ each, with one modulatory input controlling rate fluctuations in each group (See 
\figref{non-stationaryFR_simulatedData} A). In addition, we assumed that for half of each trial, there was a time-varying stimulus (`drifting grating'), represented by a 
$3$-dimensional vector which consisted of the sine and cosine of the time-varying phase of the stimulus (frequency $0.4$ Hz) as well as an additional term  which indicated whether the stimulus was on.
 
We fit Model-I N-PLDS to the data, and found that it successfully captures the non-stationarity in (log) mean firing rates, defined by $z = C(\vx + \vh) + \vd$,  as shown in \figref{non-stationaryFR_simulatedData} A, and recovers the total and trial-conditioned covariances (the across-trial mean of the single-trial covariances of $z$). For comparison, we also fit $100$ separate PLDSs to 
the data from each trial, as well as a single PLDS to the entire data.  The {{naive}} approach of fitting an individual PLDS to each trial can, in principle, follow the 
modulation. However, as each model is only fit to one trial, the parameter-estimates are very noisy since they are not well constrained by the data from each trial.
We note that a single PLDS with fixed parameters is able to track the modulations in firing rates in the posterior mean here--  however, a single PLDS would not be able to extrapolate firing rates for unseen trials (as we will demonstrate in our analyses on neural data below). In addition, it will also fail to separate `slow' and `fast' modulations into different parameters.
 By comparing the total covariance of the data (averaged across neuron pairs) to the `trial-conditioned' covariance (calculated by estimating the covariance on each trial 
individually, and averaging covariances) one can calculate how much of the cross-neuron co-variability can be explained by across-trials fluctuations in firing rates (see e.g., 
\cite{Goris_Movshon_14}). In this simulation  (which illustrates an extreme case dominated by strong across-trial effects), the conditional covariance is much smaller than the true one.

\begin{figure}[t]
\centering
\includegraphics[width=0.9\linewidth]{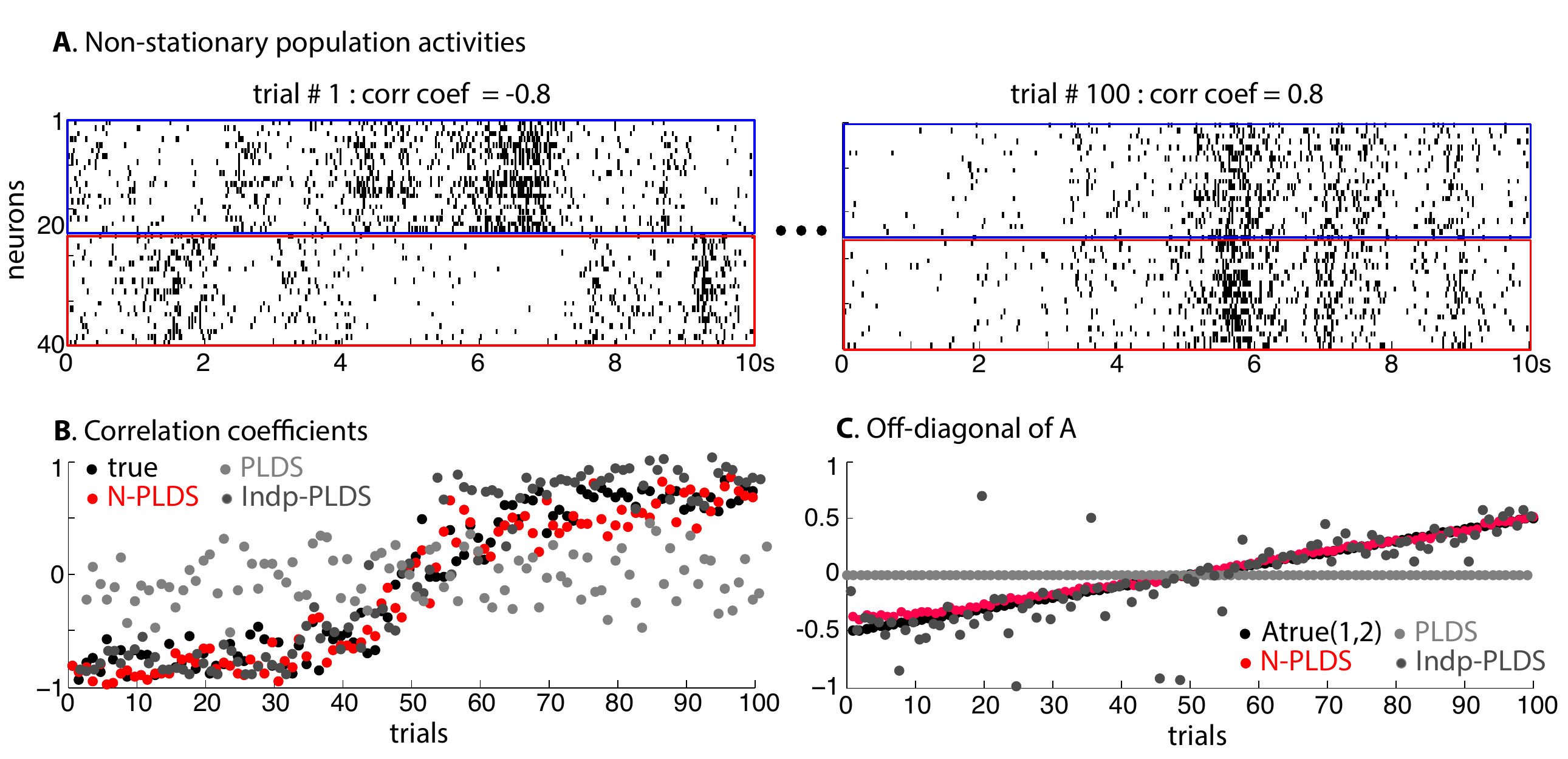} 
\caption{\textbf{Illustration of non-stationarity in population dynamics (data simulated from Model II).} \textbf{A}: Raster plots of spontaneous activity from $40$ neurons during 10 seconds of recording for 
simulated trials $1$ and $100$. We assumed that the two sub-populations (blue and red) have negative correlation at trial $1$ and positive correlation at trial $100$. \textbf{B}: 
Recovered correlations. Our Model II (red) accurately recovers the correlations between two 
groups across trials (RMSE: 0.04), while other methods perform poorly: independent PLDSs fit to each trial individually give noisy results (RMSE 0.06) and a single PLDS fit 
across all trials cannot capture the change in correlation (RMSE 0.44).  \textbf{C}: Estimation of off-diagonal in dynamics matrices. We fixed the loading matrix $C$ to its true value to avoid  issues with non-identifiability of parameters in LDS models. The off-diagonal term $A_{12}$  estimated by our model  matched the true values well, whereas the independent PLDS produced noisy estimates, and the fixed PLDS cannot capture the change in $A_{12}$.}\label{fig:non-stationaryA_simulatedData}
\end{figure}

Next, we tested Model II using a simulation of spontaneous activity from a population of $40$ neurons (simulated from Model II). We again assumed that the population could  be split into two sub-populations of size  $20$ neurons each, and simulated an experiment in which the correlation across the two sub-populations changed dramatically across  the experiment: Specifically,  we generated a $2$-d latent state that controls correlations in firing rates between the two groups of neurons, and adjusted the off-diagonal term in the 
dynamics matrix $(A_{12})$ such that the correlation between the groups varied slowly from $-1$ to $1$ across $100$ trials, where the length of each trial is $T=200$. Other elements of $A$ were adjusted such that the stationary covariance of the system was kept constant.

As before, we fit Model II N-PLDS, a single PLDS, and $100$ independent PLDSs to the data. Our model accurately recovered the correlation change in $z$ across trials, while the single  PLDS was not able to capture the non-stationarity and the independent PLDSs exhibited noisy correlations (\figref{non-stationaryA_simulatedData}). Finally,  our model also 
accurately recovered the off-diagonal parameter $A_{12}$ (\figref{non-stationaryA_simulatedData} C). For panel $C$ only, we set the loading matrix $C$ to the 
ground truth value for each of the models (Model II, fixed PLDS, separate PLDSs). LDS models suffer from non-identifiability of parameters, implying that estimated parameters do not necessarily match the true parameters even for perfect model fits. 





\begin{figure}[t]
\centering
\includegraphics[width=0.9\linewidth]{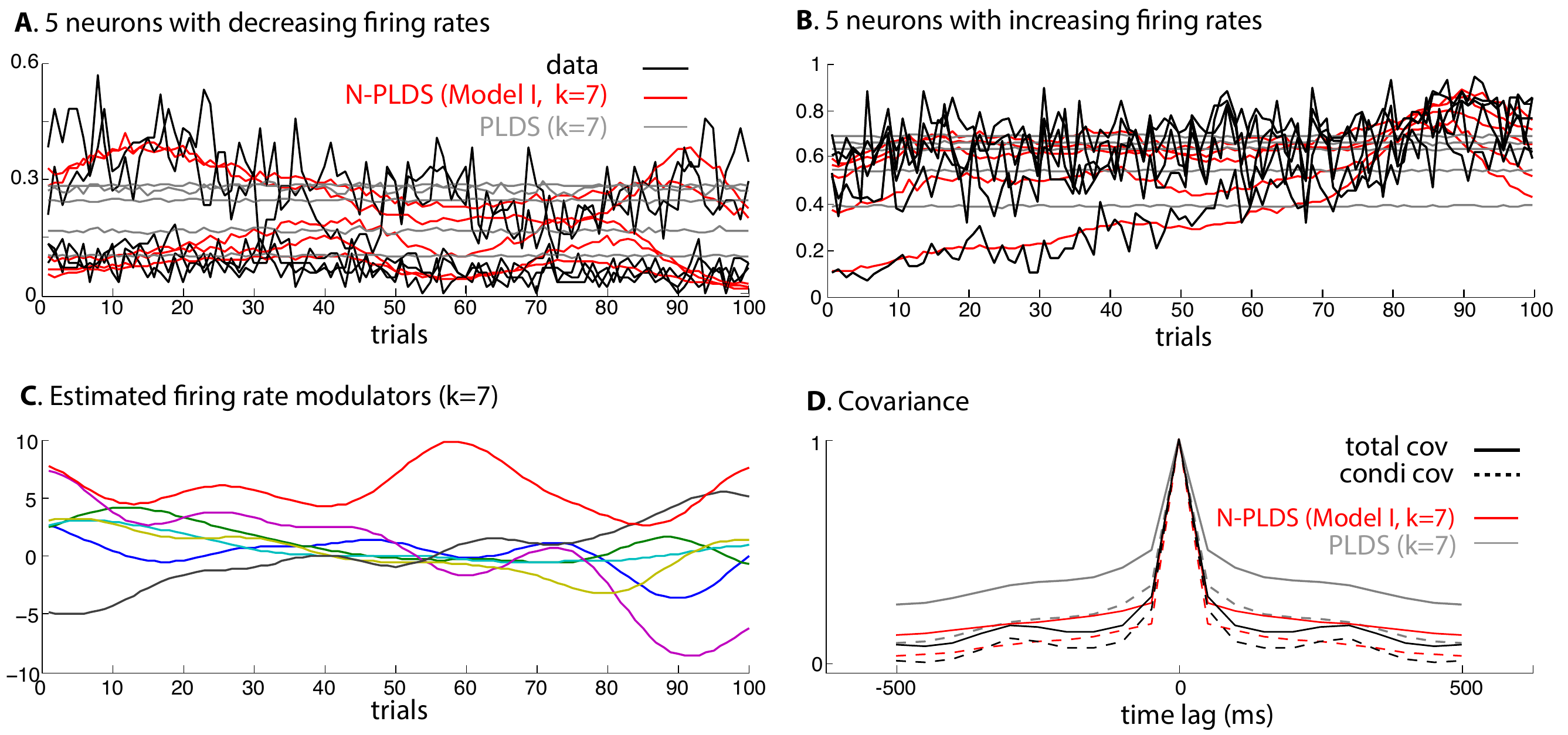} 
\caption{\textbf{Non-stationary firing rates in a population of V1 neurons.}
\textbf{A}: Firing rate change of 5 neurons (black) among $64$ with the strongest net-reduction in firing rate (quantified using linear regression on trial 
indices). \textbf{B}: Firing rate change of $5$ neurons with strongest increase in firing rates. The best RMSE on predicting mean firing rates by our method was when $k=7$, 
$RMSE=0.7383$. For comparison, single fixed-parameter PLDS across trials has an RMSE of $1.2496$. \textbf{C}: Estimated $7$-dimensional modulator $h$. The modulator 
with an estimated length scale of $\approx 10$ is smoothly varying across trials.
\textbf{D}: Comparison of total and (trial) conditional 
covariances. Our method recovers both covariances more accurately than PLDS.}
\label{fig:non-stationaryFR_realData}
\end{figure}
 
\subsection{Neurophysiological data}

\textbf{Dataset I: Non-stationary mean firing rates}

How big are non-stationarities in neural population recordings, and can our model successfully capture them? To address these questions, we first analysed a population recording from anesthetized macaque primary visual cortex consisting of  $64$ neurons stimulated by sine grating stimuli. The details of data collection are described in  \cite{Ecker_Berens_14}, but our dataset also included units that were not used in the original study due to strong non-stationarities in their firing rate.  We  binned the spikes recorded during $100$ trials of the same orientation using $50$ms bins, resulting in trials of length $T=80$ bins. Analogously to the simulated dataset above, we parameterised the stimulus as a  $3$-d vector of the sine and cosine values of the phase of the grating as well as an indicator that specifies whether there is a stimulus or not. The stimulus was presented for 2 seconds. Before and after the stimulus presentation, a blank screen was presented for 1 second. We divided the data into test data (every $10$th trial) and training data (i.e., the remaining $90$ trials).

We then fit our N-PLDS Model I to the training data. Using the  estimated parameters from the training data, we made predictions on the modulator $\vh$ on test data (we used the mean of the predictive distribution, see methods). Using these parameters, we drew samples for spikes for the entire trials 
to compute the mean firing rates of each neuron at each trial.  For comparison,  we also fit a single PLDS to the data. As this model does not allow for across-trial modulations of 
firing rates,   we simple  kept the parameters estimated from the training data.

For visualisation of results,  we used linear regression on each neuron (regression from trial index to mean firing rates) to classify neurons as `decreasing in firing rate', 
`increasing in firing rate' or `not being modulated by firing rate' based on how much of their total variance was  explained by the regression, and based on 
the sign of the regression coefficient. For the neurons with strongest modulations in firing rates, the modulations  were well captured by our model 
(\figref{non-stationaryFR_realData} A/B, results are for latent dimensionality $k=7$). For non-modulated neurons ($36$ out of $64$), our model likewise identified firing rates to 
be constant.
We tested different latent dimensions ($k=1, 5,7,9$), and found that our model works the best when $k=7$ in terms of the RMSE on mean prediction: $1.1149, 0.9398, 0.7383, 
0.7893$ for $k=1,5,7,9$, respectively.  Our model successfully captures the non-stationarity in mean firing rates (\figref{non-stationaryFR_realData} A/B) and also accurately 
recovers the total and conditional covariances (averaged across all neurons). In contrast, the fixed-parameter PLDS has firing rates which are constant across trials, therefore achieves worse reconstruction RMSE for each $k$ (e.g. RMSE: $1.2496$ for $k=7$), and also overestimates the conditional covariance.


\textbf{Dataset II : Non-stationary population dynamics}

Finally,  we analyzed a dataset of spontaneous activity recorded from a population of $40$ neurons from macaque visual cortex. The details of data collection are described in \cite{Chu2014113} and  the data is available from \cite{Figure5data}.
Using the spike-sorting information provided in the dataset, we selected the spike-cluster with highest signal-to-noise ratio from each recording channel, and out of those $46$ units kept the $40$ units with highest firing rates. As the original data consisted of one continuous recording of length $15$ minutes, we divided the data into $30$ `epochs' of length $30$ seconds each, and  used every $5$th epoch ($20\%$ of the data) for testing and the rest ($80\%$ of data) for training.

\begin{figure}[t]
\centering
\includegraphics[width=\linewidth]{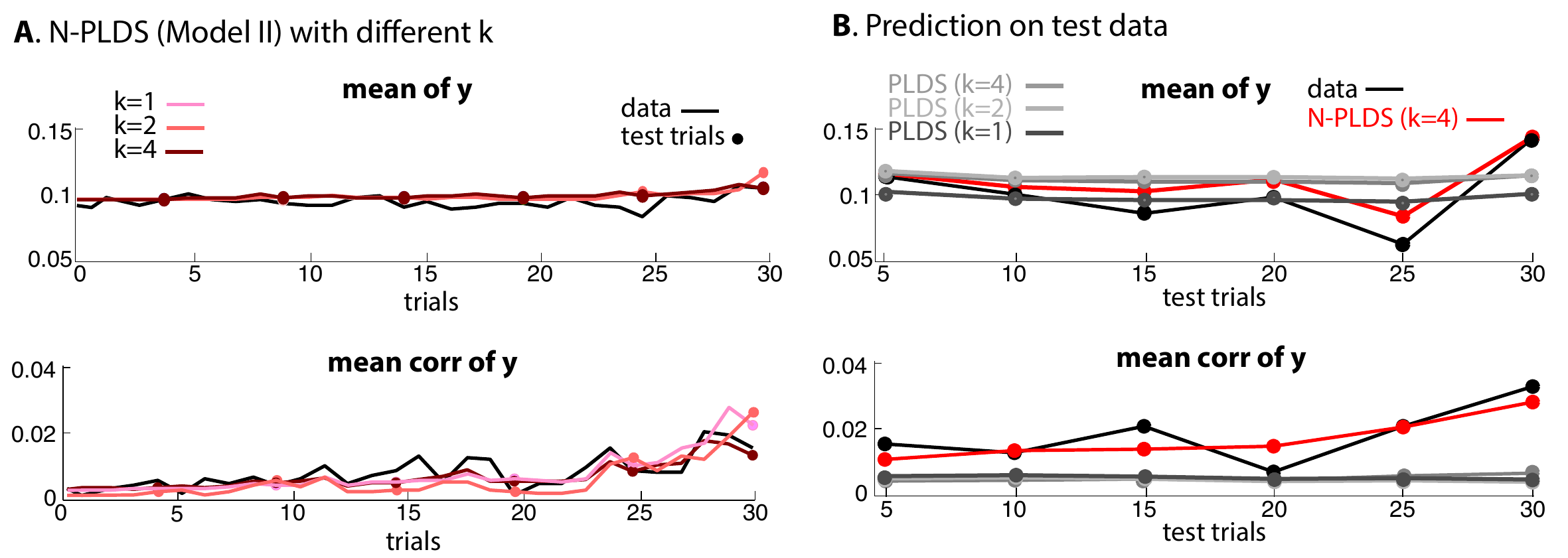} 
\caption{\textbf{Non-stationary population dynamics (data from \cite{Chu2014113})}. \textbf{A}:  Summary statistics of samples from N-PLDS (Model II) with non-stationarity 
dynamics matrix $A$ for different dimensions of latent dynamics ($k=1,2,4$). The top plot shows the mean firing rate of $40$ neurons  during $30$ epochs, showing that there 
is only a slight systematic drift in mean firing rate. Each dot represents predicted mean firing rates for the held-out data ($6$ trials). The bottom plot shows the mean correlation 
of the spike counts. All three N-PLDS models capture the increase in correlation at the end of the experiment, with the $k=4$ capturing it most accurately. \textbf{B}: Comparison 
to using a PLDS model with fixed parameters ($k=1,2,4$).  Both the mean firing rate and correlation in PLDS are constant across epochs.  As a 
consequence,  the best RMSE on mean correlation estimation in PLDS is $0.0138$ ($k=1$) compared to $0.0087$ ($k=4$) in N-PLDS.  }
\label{fig:non-stationaryA_realData}
\end{figure}

In this data, the mean firing rates are almost constant across time, while the correlations increase at the end of the experiment (\figref{non-stationaryA_realData} A). 
After estimating the parameters of our N-PLDS (Model II) from the training data, we computed the predictive distribution on the dynamics matrices $A^*$ for the test data. Using these 
parameters, we drew samples for spikes to compute the mean firing rates for each trial (\figref{non-stationaryA_realData} A), as well as the mean pairwise cross-correlations across all neuron pairs. The correlations estimated from N-PLDS (Model II) matched those in the data. For PLDS with fixed parameters, the estimated firing rates and correlations are  constant across epochs (\figref{non-stationaryA_realData} B). To quantify these results, we computed the RMSE in the prediction of mean firing rates and mean correlations on test epochs. The RMSEs on mean firing rate estimation  for PLDS are $0.0156, 0.0182, 0.0188$ for $k=1, 2, 4$, respectively, while RMSE of N-PLDS is $0.0080$ ($k=4$). The RMSE on mean correlation estimation in 
PLDSs is $0.0138$ (same for $k=1, 2, 4$) and $0.0087$ ($k=4$) in N-PLDS. 


\section{Discussion}

Non-stationarities are ubiquitous in neural data: Slow modulations in firing properties can result from diverse processes such as plasticity and learning, fluctuations in arousal, cortical reorganisation after injury as well as development and aging.  In addition, non-stationarities in neural data can also be a consequence of experimental artifacts, and can be caused by  fluctuations in anaesthesia level, stability of the physiological preparation or electrode drift. Whatever the the origins of non-stationarities are, it is important to  have statistical models which can identify them and disentangle their effects from correlations and dynamics on faster time-scales \cite{Brody_99}.  

We here presented a hierarchical Bayesian model for neural population dynamics in the presence of non-stationarity. Specifically, we concentrated on two variants of this model which focus on non-stationarity in firing rates (Model I) and dynamics matrices (Model II).   Applied to two sets of neurophysiological recording data, we demonstrated that this modelling approach can successfully capture these types of non-stationarities in neurophysiological recordings from primary visual cortex.  

There are limitations to the current study: (1) how to select amongst multiple different models which could be used to model neural non-stationarity for a given dataset; (2) how to scale up the current algorithm for larger trial numbers (e.g., using low-rank approximations to the covariance matrix); and (3) how to overcome the slow convergence properties of GP kernel parameter estimation \cite{NIPS2010_0835}. 
We believe that extending our method to address these questions provides an exciting direction for future research,  and will result in a powerful set of statistical methods for investigating how neural systems operate in the presence of non-stationarity.

\section*{Acknowledgments} 
We thank the authors of \cite{Ecker_Berens_14} for sharing dataset I with us and making it publicly available at: http://toliaslab.org/publications/ecker-et-al-2014/, as well as the authors of \cite{Chu2014113} for making dataset II publicly available via CRCNS.org.
We thank Srinivas Turaga, Thomas Desautels, Alexander Ecker, and Maneesh Sahani for their valuable comments on the manuscript. This work was funded by the German Federal Ministry of Education and Research (BMBF; FKZ: 01GQ1002, Bernstein Center T{\"u}bingen) and the Gatsby Charitable Trust.

\footnotesize
\bibliographystyle{unsrt} 

\bibliography{nips2014}

\end{document}